\documentclass[11pt]{article}

\usepackage[final]{acl}

\usepackage{times}
\usepackage{latexsym}

\usepackage[T1]{fontenc}

\usepackage[utf8]{inputenc}

\usepackage{microtype}

\usepackage{inconsolata}

\usepackage{graphicx}
\usepackage{listings}
\usepackage{xcolor}
\usepackage{multirow}
\usepackage[normalem]{ulem}
\useunder{\uline}{\ul}{}
\usepackage{booktabs}
\usepackage{multirow}
\usepackage{colortbl}
\usepackage{amssymb}
\usepackage[normalem]{ulem}
\usepackage{enumitem} 
\usepackage[accsupp]{axessibility}
\usepackage[dvipsnames]{xcolor}
\definecolor{upgreen}{HTML}{1A9641}
\usepackage{cleveref}

\newcommand{\dataset}{\textsc{Affordance20Q}}
\title{\dataset: Evaluating Affordance Reasoning from Physical Properties}

\author{Yifan Jiang$^{1}$,
Meige Yang$^{2}$, Zitong Li$^{2}$, Jay Pujara$^{1}$ \\ 
$^1$Information Sciences Institute, University of Southern California \\
 $^2$University of Southern California\\
  \texttt{\{yifjia,jpujara\}@isi.edu, maggieya@usc.edu, alex.zitong.li@gmail.com}}

\begin{document}
\maketitle
\begin{abstract}

Affordance reasoning, the inference of an object's action
possibilities from its physical properties~(e.g., shape and material),
is fundamental to human physical understanding and increasingly
critical for Large Language Models~(LLMs). However, existing
affordance benchmarks largely expose explicit object identities in the
evaluation setup, allowing models to rely on memorized
object-affordance mappings rather than reasoning over physical
properties. To address this gap, we introduce \dataset, a novel
affordance reasoning benchmark formulated as a 20-Questions game
without exposing the object's identity. In each game, the model
identifies a hidden object's affordance from a candidate set by asking
yes/no questions about its physical properties. \dataset~comprises
1{,}009 games over 454 objects and 59 affordances, all manually
filtered, refined, and annotated. We conduct comprehensive experiments
with 15 state-of-the-art LLMs and find a substantial gap~($\sim$20 points) compared to human
performance. A KL-based information-gain~(IG) analysis further shows that models fail to ask discriminating questions as the game progresses. To
close the gap, we develop KB-Anchored Rule Induction~(KARI), a
pipeline based on LLMs that generates affordance rules grounded in evidence from knowledge bases~(KBs). KARI improves open-source LLMs by up to 15.2 points, while the limited coverage of KBs hinders further gains. We release all our code and data at \url{https://github.com/1171-jpg/Affordance20Q.git}.

\end{abstract}

  \begin{figure}[t]                                         
    \centering
    \includegraphics[width=1\linewidth]{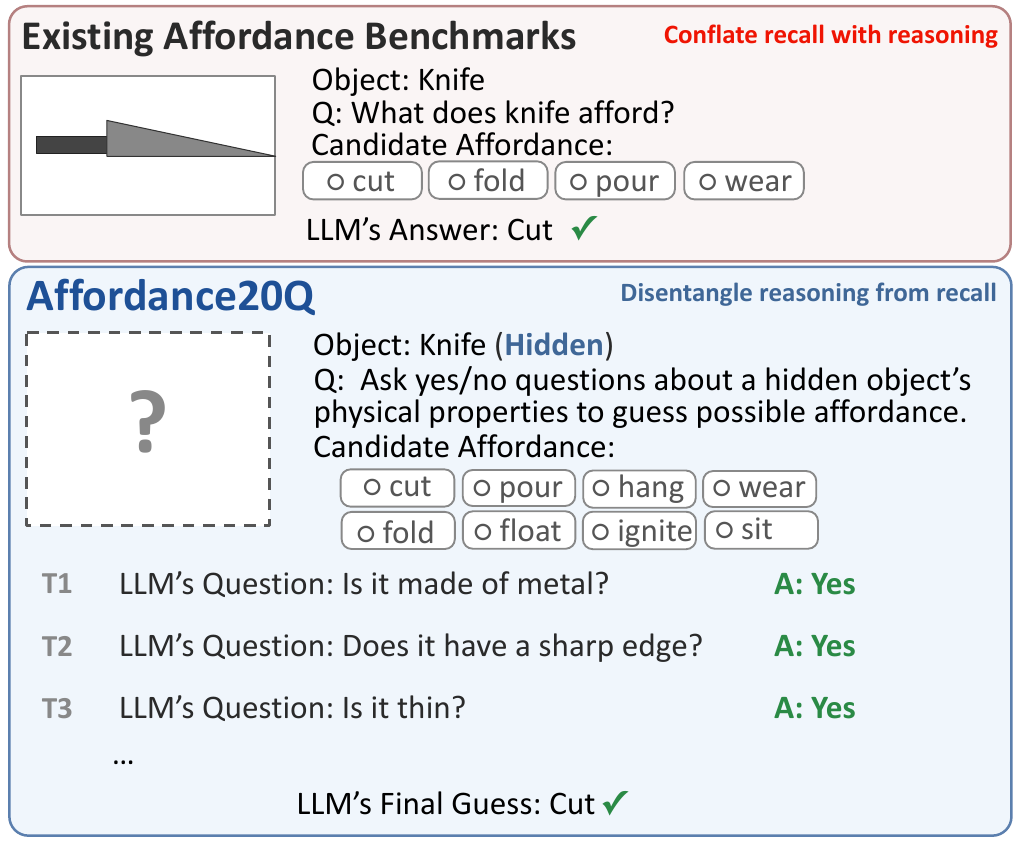}
    \caption{Comparison between existing affordance benchmarks~(top) and \dataset~(bottom).}                                                                      
    \label{fig:main_example}                                                                    
  \end{figure}   

\section{Introduction}
When humans encounter an object, they reason about what actions it
supports directly from its physical properties, including its shape,
material, and structure. \citet{gibson1977theory} formalized this
capacity as \textbf{affordance reasoning} and placed it at the core
of physical understanding. Affordance reasoning operates on physical
properties that every object exhibits, and therefore supports a wide
range of physical interaction, from fluently using familiar everyday
objects~\cite{norman2013design} to robustly handle novel ones or
creatively repurposing familiar ones beyond their typical
use~\cite{duncker1945problem,german2000immunity}. Such reasoning is increasingly crucial for Large Language
Models~(LLMs) as recent progress has led to their integration into
daily human activities~\cite{yang2025qwen3,singh2025openai}, and
especially for the growing class of LLM-driven embodied
systems~\cite{driess2023palm,zhang2025landscape} that
aim for physical application. LLMs are expected not only to
understand everyday tool
use~\cite{wang2023newton,wang2026affordance}, but also to generalize
this capability to unfamiliar objects or scenarios beyond their
training experience~\cite{tian2024macgyver,jiang2023brainteaser}. For example, in embodied robotic manipulation, a model should not only recognize that a knife affords cutting, but also reason that any object with a sharp rigid edge affords the same action~\cite{tang2025affordgrasp,xu2022partafford}.

Recognizing its importance, recent research has introduced diverse
benchmarks for affordance reasoning, spanning multiple
modalities~\cite{wang2026affordance,yu2025seqafford} and task
types~\cite{qasemi2022paco,tian2024macgyver}. Yet all
setups make the object's identity explicit by giving the class or
object name~(\Cref{fig:main_example}), conflating recall of object-affordance mappings with
reasoning from physical properties. A model knowing that the object
is a knife can answer ``it affords cutting'' simply by recalling a
stored mapping~\cite{persiani2019unsupervised}, without ever
consulting its rigid, sharp-edged physical properties. Such conflation prevents existing benchmarks from accurately measuring
affordance reasoning capability. Recall-based models can pass these
benchmarks within the training distribution, but fail on the unseen
objects and atypical uses common in real-world physical
interaction~\cite{gjerde2025reasoning,wu2024reasoning}.

To disentangle reasoning from recall, we introduce \dataset, a
benchmark cast in a 20-Questions~(20Q) game
setting~\cite{bruner1966studies}, in which models identify an
object's affordance without knowing its identity. As shown in
\Cref{fig:main_example}, in each game, given candidate affordances, the model narrows the candidates through multi-turn yes/no
questions about the hidden object's physical properties, such as
material and shape. The exclusion of object identity requires the model to identify the correct 
affordance by reasoning from physical evidence rather than
recalling stored object-affordance
mappings~\cite{hutson2025guessinggame}. To construct
\dataset, we use a three-stage pipeline that first collects physical
objects and affordances from the existing corpus~\cite{jiang2021learning} and commonsense knowledge
base~\cite{ilievski2021cskg}, then enriches them via LLM generation,
and finally manually filters, refines, and annotates (object,
affordance) pairs. In the end, \dataset~consists of 1{,}009 games covering 454 objects and 59 affordances.


With \dataset, we evaluate 15 state-of-the-art LLMs with different sizes and architectures, and find a substantial gap~($\sim$20 points) compared to human performance, with the strongest model reaching only 45.9\%. To further diagnose the gap, we trace each game with a KL-based information-gain~(IG) metric and find that models repeatedly ask low-IG questions that fail to narrow the candidate set across turns. To close this gap, we develop KB-Anchored Rule Induction~(KARI), a pipeline that uses knowledge bases to both inspire LLM rule generation and re-ground the generated rules through post-hoc validation, ensuring that generated rules remain anchored in physical commonsense rather than free-form LLM speculation. KARI improves open-source LLMs by up to 15.2 points, partially closing the gap, while the remaining shortfall traces to the coverage limits of current commonsense knowledge bases.

We summarize our contributions as follows: 1) We introduce \dataset, a 20-Questions benchmark that tests affordance reasoning from physical properties rather than object-identity recall, comprising 1{,}009 games over 454 objects and 59 affordances. 2) We conduct comprehensive experiments with 15 state-of-the-art LLMs, revealing a substantial gap to human performance, with information-gain analysis showing that models fail to ask discriminating questions as the game progresses. 3) We develop KB-Anchored Rule Induction~(KARI), a pipeline that combines LLMs and knowledge bases, improving open-source LLMs by up to 15.2 points. We release all code and data.

\section{Related Work}

\paragraph{Affordance Reasoning Benchmarks} The importance of affordance reasoning~\cite{gibson1977theory} has motivated a wide range of benchmarks across diverse input formats and modalities. In the vision domain, early work focuses on grounding part/object-level affordance in either 3D shape and part geometry~\cite{deng20213d,xu2022partafford} or 2D image input~\cite{nguyen2017object,luo2022learning,li2023beyond}. Recent work shifts to evaluating models' affordance reasoning with both image inputs and text instructions across different task setups~\cite{wang2026affordance,yu2025seqafford,wang2026affordancer1,zhu2025afford,wan2025instructpart}. In the text domain, parallel work formats affordance reasoning as question-answering tasks with object names or descriptions provided as context~\cite{bisk2020piqa,aroca-ouellette-etal-2021-prost,wang2023newton,adak2024text2afford,gjerde2025reasoning}. Although a few works~\cite{li2023beyond,gjerde2025reasoning} format affordance reasoning based on physical properties, none of them remove object identity to prevent recall of object-affordance mappings. In contrast, \dataset~is the first benchmark to exclude object identity from the input and require the model to infer affordance from physical properties through multi-turn questioning, disentangling reasoning from recall.

\paragraph{20-Questions Games and Active Question-Asking}
The 20-Questions~(20Q) game was first used in cognitive science to study information-seeking behavior~\cite{bruner1966studies,ruggeri2016sources}.
In each game, a questioner aims to identify a hidden target through a sequence of yes/no questions within a fixed number of turns.
As active question-asking ability becomes increasingly important in real-world human-computer interaction scenarios (e.g., task disambiguation~\cite{kobalczyk2025active}, medical diagnosis~\cite{li2024mediq}), the 20Q game has recently been widely adopted to analyze LLMs' multi-turn reasoning and information-seeking abilities~\cite{bertolazzi2023chatgpt,hutson2025guessinggame,zhang2024probing,mazzaccara-etal-2024-learning}.
However, all current work chooses the object or entity as the candidate space.
For example, \citet{zhang2024probing} and \citet{hutson2025guessinggame} require LLMs to strategically ask questions to identify a hidden target object.
\dataset~is the first to adopt the 20Q game for affordance reasoning, which also aligns with Gibson's active-perception view of affordance~\cite{gibson1977theory}.
We further introduce a novel KL-based information-gain metric to evaluate model question effectiveness across turns.
\section{\dataset~Construction}
\label{sec:dataset}
In this section, we first formalize the game setup of \dataset~(\S\ref{sec:dataset:task}), then describe the design of our three-stage collection pipeline~(\S\ref{sec:dataset:pipeline}), and end up with the data statistics~(\S\ref{sec:dataset:statistic}).
\subsection{Game Formulation}
\label{sec:dataset:task}

Following recent 20Q adaptations for LLM evaluation~\cite{hutson2025guessinggame,zhang2024probing}, each game in \dataset~is defined by a hidden target object $o^*$ and a candidate affordance set $\mathcal{A} = \{a^*, a_1, \dots, a_7\}$ of 8 affordances, in which the target $a^*$ is an affordance that $o^*$ possesses and the remaining 7 are distractors. Three agents participate: a \textit{Questioner} that observes only $\mathcal{A}$ and asks yes/no questions~($q$) about $o^*$'s physical properties to identify $a^*$, a \textit{Checker} that ensures each question is both well-formed and grounded in one of the physical-property dimensions (e.g., material, shape), preventing information leakage, and an \textit{Oracle} that has access to $o^*$ and provides answers~($r$) to physical-property questions. At each turn $t$, the Questioner produces a question $q_t$ based on the dialogue history $H_{t-1} = \{(q_1, r_1), \dots, (q_{t-1}, r_{t-1})\}$, the Checker validates $q_t$, and the Oracle returns an answer $r_t$. The game succeeds if the Questioner correctly identifies $a^*$ within a budget of $T=20$ turns. It fails if the Questioner makes an incorrect guess or exhausts the turn budget.

\subsection{Three-Stage Collection Pipeline}
\label{sec:dataset:pipeline}
Unlike previous 20Q setups~\cite{hutson2025guessinggame,zhang2024probing} that frame the task as identifying a hidden object, to evaluate affordance reasoning, \dataset~frames it as identifying a hidden object's affordance, where the hidden object remains but the Questioner reasons over its physical properties. To ensure this reasoning chain is valid, \dataset~excludes any object-affordance pair whose affordance is not deducible from physical-property dimensions alone~(e.g., a microwave's heating affordance comes from its magnetron). Given the cost of fully manual curation, we design a semi-automatic three-stage collection pipeline to construct \dataset~in a scalable way.

\noindent\textbf{Stage 1: Initial Object and Affordance Collection.} Our initial object pool is the list of human-made physical objects introduced in \cite{jiang2021learning}. We build the initial affordance pool by querying Commonsense Knowledge Graph~(CSKG)~\cite{ilievski2021cskg} for the relations describing what objects are capable of and used for (e.g., \texttt{CapableOf}, \texttt{UsedFor}). CSKG integrates seven commonsense knowledge bases (e.g., ConceptNet~\cite{speer2017conceptnet}, WordNet~\cite{fellbaum2010wordnet}) into one schema and thus offers denser affordance coverage than any single source.\\
\noindent\textbf{Stage 2: Affordance Expansion and Pool Filtering.} To further expand the affordance coverage, we query CSKG for each object's physical properties and prompt an LLM to propose additional candidate affordances conditioned on them. To remove entries unsuitable for our task, we then manually filter both pools based on specific principles. For affordances, we remove entries that violate the following three dimensions: (1) \textit{context-dependence}, the affordance depends on external context rather than the object itself~(e.g., being a gift), (2) \textit{mechanism-dependence}, the affordance depends on hidden internal mechanisms~(e.g., playing video), and (3) \textit{over-generality}, the affordance is too abstract to be discriminative~(e.g., useful). For objects, we drop entries that are not discrete physical objects~(e.g., raw material, human body part) or whose function depends on hidden internal mechanisms~(e.g., microwave, TV). After filtering, we refine affordance names and write a short description for each, and consolidate each object's CSKG properties by removing irrelevant or conflicting entries, forming each object's \textit{property set} for Stage 3~(Details in \Cref{app:annotation}).\\
\noindent\textbf{Stage 3: Annotation and Object Specialization.} In this stage, we annotate every (object, affordance) pair, determining whether each affordance applies to its paired object. Additionally, as each object's \textit{property set} from CSKG tends to be generic and lacks specific physical details (e.g., some spatulas have an internal hollow bar that filters water), we also specialize each object's property set during annotation. We first prompt an LLM to label every~(object, affordance) pair as \textit{YES}, \textit{NO}, or \textit{MAYBE} given the object's property set and the affordance's definition, naming the additional property required for each \textit{MAYBE} label. We then manually verify each \textit{YES}/\textit{NO} label against the object's property set, and for each \textit{MAYBE}, decide whether the proposed property applies, and if it does, add the property to the set, turn the label to \textit{YES}, and re-check consistency.

\subsection{Implementation Details and Data Statistics}
\label{sec:dataset:statistic}
We run the pipeline with GPT-4.1~\cite{achiam2023gpt} as the main LLM due to its promising performance in public benchmarks~\cite{liang2022holistic}. Six human annotators were involved in the manual annotation and refinement process in Stages 2 and 3. We further recruited three additional annotators to re-annotate a random subset of 1{,}298 pairs, achieving an 85.2\% majority agreement with our released labels (Fleiss $\kappa= 0.82$~\cite{fleiss1971measuring}). Among the three annotators themselves, pairwise agreement averages 90.9\%, and all three agree on 86.4\% of the pairs. Applying the three-stage pipeline yields 454 objects, each with a property set, 59 affordances, and a label for every (object, affordance) pair. We sample 1{,}009 (object, target affordance) game instances as the test set, balanced across target affordances so that no single affordance dominates, forming \dataset.

\section{KB-Anchored Rule Induction}
\label{sec:method}

Prior work on affordance and commonsense reasoning either distills rules from KBs bound by a fixed schema~\cite{zhu2014reasoning}, or generates them with LLMs but risks hallucination~\cite{west2022symbolic}. We investigate whether combining KB knowledge with LLM reasoning yields rules that are both grounded and expressive, thereby assisting the Questioner in our task. Therefore, we develop KB-Anchored Rule Induction~(KARI), a pipeline that uses an LLM to induce a compositional rule per affordance grounded in evidence from external KBs. We next describe KARI's rule format and how its components combine KB knowledge with LLM reasoning during rule generation.

  \begin{figure}[t]                                         
    \centering
    \includegraphics[width=1\linewidth]{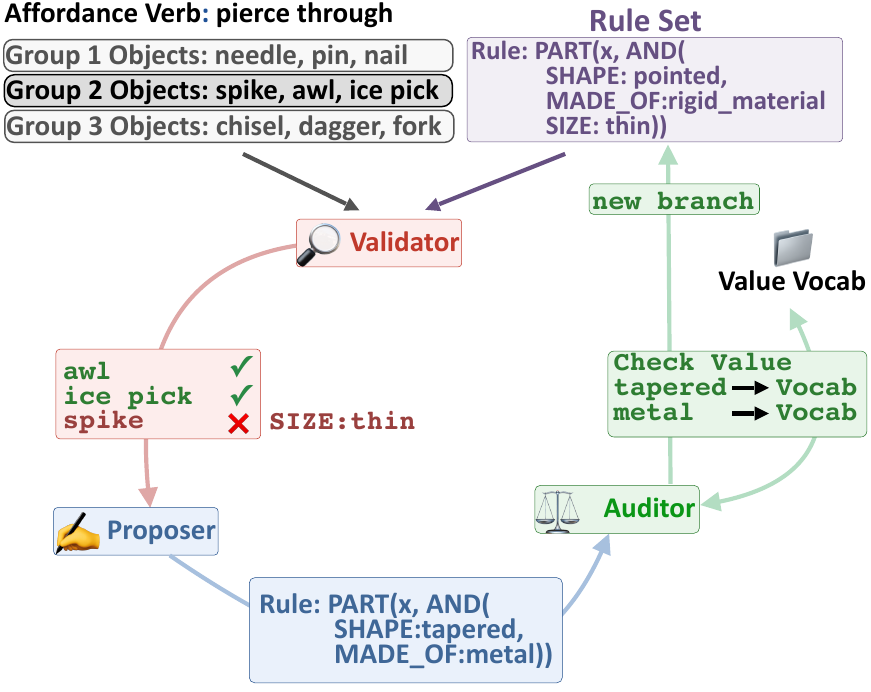}
    \caption{Illustration of KB-Anchored Rule Induction~(KARI)}                                                                      
    \label{fig:kari_pipeline}                                                                    
  \end{figure} 

\subsection{Rule Format}
A KARI rule is a tree expression stating the static physical conditions enabling an affordance. KARI's rule grammar has seven operators in three categories. (i) Four atomic \textbf{predicates} describe a physical property of the entity: \texttt{MADE\_OF}, \texttt{SHAPE}, \texttt{SIZE}, and \texttt{SURFACE}. Each takes a \textbf{value} to represent one physical property. For instance, \texttt{MADE\_OF: metal} asserts that the entity is made of metal. (ii) Two \textbf{combinators} produce a \textbf{combination} of multiple atomic predicates: \texttt{AND} and \texttt{OR}, which can nest inside each other. For instance, \texttt{AND(MADE\_OF: metal, SHAPE: pointed)} requires the entity to be both metal and pointed.  (iii) The constructor \texttt{PART(x, predicate/combination)} represents a specific part of the object, where \texttt{x} is a variable that can refer to any part. We introduce \texttt{PART} because many affordances can be enabled by one specific part rather than the entire object, and we use additional part variables (e.g., \texttt{y}, \texttt{z}) when multiple distinct parts are required.

\subsection{Rule Proposer, Validator, and Auditor}
KARI consists of three components, each for a specific purpose~(\Cref{fig:kari_pipeline}). We first introduce each component, then illustrate the overall KARI pipeline. We collect affordance verbs\footnote{We use \textit{affordance verb} to refer to affordances collected from KBs in the KARI pipeline.} and corresponding positive objects together with their materials, parts, and other physical properties from CSKG~\cite{ilievski2021cskg} and the Aristo Tuple KB~\cite{dalvi2017domain} as our initial data.

\noindent\textbf{Proposer.} The proposer is the only component that writes rule expressions. Given an affordance verb and a set of objects with their physical properties from a knowledge base, it either drafts a new rule or extends an existing one by appending a new \texttt{OR} branch. We restrict the proposer's behavior to be add-only to ensure possible errors will only occur in the new branch.

\noindent\textbf{Validator.} The validator takes an object and a set of rules as input. Since the object is a positive example of the affordance verb according to the KB, each rule is expected to evaluate true on it. For each rule, an LLM labels every atom as \textit{YES} or \textit{NO}, and a code script evaluates the \texttt{AND}/\texttt{OR} tree to determine whether the rule covers the object.

\noindent\textbf{Auditor.} The auditor validates each value the proposer introduces and maintains a vocabulary of accepted values per predicate, which prevents ill-typed values from entering the rule and the vocabulary from drifting through redundant synonyms. Given a rule, the auditor checks whether each value is valid for its predicate~(e.g., \texttt{MADE\_OF: cylinder} is rejected), then either maps the value to an existing synonym in the vocabulary or adds it as a new entry for future use.

\subsection{Pipeline}
\Cref{fig:kari_pipeline} shows the overview of the pipeline. We split the collected data into three groups and call each component in a loop. In round~1, the proposer drafts a seed rule on the first group of data, and the auditor updates the values and the vocabulary. In rounds~2 and~3, the validator first validates the previous rule, and the proposer proposes a new rule for uncovered objects in the new group, with the auditor performing the same update again. 

Adapting KARI to our collected data with \textsc{Qwen3-14B} for all components produces 2{,}223 rules. In the inference time, for each game, we compute the sentence similarity~\cite{ni2022sentence} between each candidate affordance and every generated rule's affordance verb, and match the candidate to its highest-scoring rule when the similarity exceeds 0.7. Matched rules are verbalized at the end of the Questioner's system prompt. A candidate affordance with no match above the threshold receives no rule. Full details and rule examples are presented in \Cref{app:kari_pipe}.
\section{Experiment Setup}
\label{sec:setup}

We evaluate on the 1{,}009-game test split of \dataset~(\Cref{sec:dataset}), with the human baseline run on a sampled subset.
The Questioner, Oracle, and Checker all run at temperature 0. Full prompts and other details are in \Cref{app:setup}.

\subsection{Questioners}

To ensure a comprehensive evaluation, we test 15 LLMs as Questioners, grouped into open-source and closed-source models.

\paragraph{Open-source LLMs.} We evaluate ten open-source models spanning a wide range of scales. Eight are dense models between 8B and 14B parameters: \textsc{Qwen3-8/14B} and \textsc{Qwen3.5-9B}~\cite{yang2025qwen3}, \textsc{Phi-4-14B}~\cite{abdin2024phi}, \textsc{Llama-3.1-8B}~\cite{grattafiori2024llama}, \textsc{Ministral-8B}~\cite{liu2026ministral}, \textsc{Nemotron-9B}~\cite{blakeman2025nvidia}, and \textsc{Gemma-3-12B}~\cite{gemmateam2025gemma3technicalreport}. The other two are the mixture-of-experts models \textsc{DeepSeek-V4-Pro} and \textsc{DeepSeek-V4-Flash}~\cite{deepseek2026v4}.

\paragraph{Closed-source LLMs.} We evaluate five proprietary models accessed through their official APIs: \textsc{GPT-5} and \textsc{GPT-5-mini}~\cite{singh2025openai}, \textsc{Gemini-2.5-Pro} and \textsc{Gemini-2.5-Flash}~\cite{comanici2025gemini}, and \textsc{MiniMax-M2.5}~\cite{minimax2025m2}.

\paragraph{Reference points.} 1)~\textsc{\textbf{Human}}: five annotators play a 30\% subset under the same game rules, and we report the average result. We further include two non-LLM reference points to show the possible performance range. To enable the final guess, for each candidate affordance, we select ten representative objects, and use the fraction of them that remain consistent with all questions and their corresponding Oracle answers. We then softmax these fractions across affordances as the likelihood of being the target. 2)~\textsc{\textbf{Fix20Q}} asks the 20 most frequent questions from the LLM evaluation in a fixed order (Appendix~\ref{app:fixedq}), then samples its final guess from a softmax over these likelihoods, and we report the average over five runs. 3)~\textsc{\textbf{Optimal}} is an information-theoretic upper bound with full access to all candidates' representative objects. At every turn, it asks the question that drives the target affordance's likelihood as high as possible, and gives the final answer once a single affordance's likelihood exceeds a specific threshold~(0.9).

We also test \textsc{Entropy-Max}, which asks the question that maximally reduces the entropy over the eight candidates without knowing the target. It reaches 51.2\% in 4.6 turns, showing that narrowing the search space alone is not enough for success (details in \Cref{app:entropy}).

\subsection{Oracle and Checker}

We use a single \textsc{Qwen3-14B} instance for both the Oracle and the Checker. Since we provide each object's \textit{property set}~(\Cref{sec:dataset:pipeline}) as context, the Oracle agrees with human labels on 93.4\% of 300 manually verified questions (MATERIAL 98.0\%, SHAPE 95.7\%, SURFACE 88.2\%, SIZE 84.8\%), with errors concentrated in borderline size comparisons~(\Cref{app:oracle}). We also test a stronger \textsc{GPT-5} Oracle on six Questioners, which shifts success rates by at most 2.6 points and leaves the ranking unchanged~(\Cref{app:oracle}).

\subsection{Metrics}

We report three metrics: 1) \textbf{Success Rate~(SR)}, the fraction of games in which the target affordance is identified within the 20-turn budget, 2) \textbf{Turns}, the average number of turns in solved games, and 3) \textbf{Information Gain~(IG)}, a per-turn measure of how fast each question narrows the candidate affordances. Using the same representative objects, let $n_t(a)$ be the number of affordance $a$'s representative objects still consistent with the dialogue history after turn $t$, giving a distribution $b_t(a) = n_t(a) / \sum_{a'} n_t(a')$ over the 8 candidates. IG is the KL divergence between consecutive distributions,
\begin{equation}
\mathrm{IG}_t = D_{\mathrm{KL}}\!\left(b_t \,\|\, b_{t-1}\right) = \sum_{a} b_t(a)\,\log_2 \frac{b_t(a)}{b_{t-1}(a)}.
\end{equation}
\begin{table*}[!ht]
\centering
\footnotesize
\renewcommand{\arraystretch}{0.9}
\setlength{\tabcolsep}{10pt}
\begin{tabular}{lcccc}
\toprule
& \multicolumn{2}{c}{Vanilla} & \multicolumn{2}{c}{+KARI} \\
\cmidrule(lr){2-3}\cmidrule(lr){4-5}
Questioner & SR & Turns & SR & Turns \\
\midrule
\multicolumn{5}{l}{\textit{Reference Points}} \\
\textsc{Optimal}             & 100.0 &  2.5 & -- & -- \\
\textsc{Human}               &  64.2$\pm$7.3 & 10.7$\pm$1.3 & -- & -- \\
\textsc{Fix20Q}              &  24.8$\pm$0.9 & 20.0 & -- & -- \\
\midrule
\multicolumn{5}{l}{\textit{Open-source LLMs}} \\
\textsc{Llama-3.1-8B}        & 18.4 & 14.5 & 31.3\textsubscript{\color{upgreen}$\uparrow$12.9} & 20.0 \\
\textsc{Ministral-8B}        & 17.7 & 15.1 & 28.0\textsubscript{\color{upgreen}$\uparrow$10.3} & 12.6 \\
\textsc{Qwen3-8B}            & 26.3 & 16.1 & 33.4\textsubscript{\color{upgreen}$\uparrow$7.1}  & 11.7 \\
\textsc{Nemotron-9B}         & 14.9 &  8.6 & 30.0\textsubscript{\color{upgreen}$\uparrow$15.1} & 17.4 \\
\textsc{Qwen3.5-9B}          & 25.1 & 18.6 & 33.2\textsubscript{\color{upgreen}$\uparrow$8.1}  & 14.3 \\
\textsc{Gemma-3-12B}         & 27.4 & 15.1 & 33.7\textsubscript{\color{upgreen}$\uparrow$6.3}  & 18.6 \\
\textsc{Qwen3-14B}           & 27.3 & 18.6 & 32.9\textsubscript{\color{upgreen}$\uparrow$5.6}  & 10.9 \\
\textsc{Phi-4-14B}           & 26.4 & 11.2 & 30.5\textsubscript{\color{upgreen}$\uparrow$4.1}  & 11.9 \\
\textsc{DeepSeek-V4-Flash}   & \underline{41.3} & 16.9 & 37.9\textsubscript{\color{red}$\downarrow$3.4}    & 10.6 \\
\textsc{DeepSeek-V4-Pro}     & 40.5 & 18.8 & \underline{38.5}\textsubscript{\color{red}$\downarrow$2.0}    & 11.2 \\
\midrule
\multicolumn{5}{l}{\textit{Closed-source LLMs}} \\
\textsc{GPT-5-mini}          & 22.8 & 16.9 & 31.2\textsubscript{\color{upgreen}$\uparrow$8.4}  & 10.9 \\
\textsc{GPT-5}               & 35.5 & 14.5 & 32.4\textsubscript{\color{red}$\downarrow$3.1}    &  7.0 \\
\textsc{MiniMax-M2.5}        & 34.9 & 17.2 & 35.5\textsubscript{\color{upgreen}$\uparrow$0.6}  &  9.3 \\
\textsc{Gemini-2.5-Flash}    & 41.4 & 17.3 & 33.4\textsubscript{\color{red}$\downarrow$8.0}    &  9.9 \\
\textsc{Gemini-2.5-Pro}      & \textbf{\underline{45.9}} & 18.5 & \textbf{\underline{40.0}}\textsubscript{\color{red}$\downarrow$5.9}    & 12.1 \\
\bottomrule
\end{tabular}
\caption{Main results on \dataset, with and without KARI's rule integration. SR is the success rate (\%), and Turns is the average turns. The best overall result is in \textbf{bold}, the best in each category \underline{underlined}.}
\label{tab:main}
\end{table*}

\section{Results}
\label{sec:results}

We focus on five research questions: \textit{1) How well do LLMs reason about affordances from physical properties? 2) How do LLMs behave when reasoning about 
different affordances? 3) How effectively do LLMs gather information through questioning? 4) Can KARI's rules improve LLMs' affordance reasoning? 5) What are the typical success and failure modes?}

\subsection{Main Results}
\label{sec:results:main}
We report the success rate and average turns in \Cref{tab:main}. \textsc{Optimal} reaches 100\% in 2.5 turns, while \textsc{Fix20Q} reaches only 24.8\% despite using the full 20-question budget, showing that extensive questioning without affordance reasoning specific to each game is not beneficial for success. Humans reach 64.2\% in 10.7 turns, and their gap to \textsc{Optimal} shows the challenging nature of
\dataset. We note that 2.5 turns is a theoretical minimum, achievable only by optimally ordering questions to rule out distractor affordances or confirm the target affordance, whereas humans can ultimately solve the games in more turns, confirming the validity of \dataset.

For both open- and closed-source LLMs, \textbf{all models fall far short of human performance, with gaps ranging from roughly 20 to 50 points}. Most open-source LLMs perform similarly to \textsc{Fix20Q}, ranging from 14.9\% to 27.4\%, showing they struggle to solve each game with appropriate affordance reasoning. \textsc{DeepSeek-V4-Flash} and \textsc{DeepSeek-V4-Pro} are the clear exceptions, consistent with their strong results on other benchmarks~\cite{deepseek2026v4} and likely aided by their mixture-of-experts architecture. All closed-source LLMs except \textsc{GPT-5-mini} achieve higher success rates than the 8B to 14B models. Notably, \textsc{Gemini-2.5-Pro} outperforms the strongest open-source LLM~(\textsc{DeepSeek-V4-Flash}) by 4.6 points, showing a 4.6-point gap between the two groups' best models. Models also differ in their turn usage, which does not guarantee a higher success rate. For example, \textsc{Nemotron-9B} and \textsc{Phi-4-14B} tend to end games early, while \textsc{Qwen3.5-9B} tends to exhaust its turn budget, but neither yields a higher success rate.

\subsection{Affordance Difficulty}
\label{sec:results:diff}
We further break down model performance by affordance (full results in \Cref{app:sr_breakdown}), which ranges widely from 3.1\% to 56.0\%. In general, affordances that can be inferred from a single physical property tend to be easy, such as \textit{transmit\_light}~(51.4\%), established once the object is known to be transparent, or \textit{conduct\_heat}~(55.2\%), once it is known to be made of metal material. The hardest affordances instead require reasoning over multiple physical properties or alternative affordance rules, such as \textit{sink\_in\_water} (3.8\%), which jointly depends on the material of the object, whether it contains a hollow structure inside, and whether the material can absorb the water, or \textit{hang\_from\_above} (10.6\%), where the object has a hook-shaped, ring-shaped, or strap-like part that can support this affordance. We next compare open- and closed-source LLMs' performance on each affordance. Open-source LLMs often perform better when an affordance hinges on a single physical property, while closed-source LLMs spend more turns yet result in a lower success rate, suggesting a potential overthinking behavior. On the other hand, closed-source LLMs outperform open-source LLMs on affordances requiring multiple physical properties, with similar turn usage, highlighting their stronger reasoning ability.

\label{sec:results:ig}
\begin{figure}[h]
\centering
\includegraphics[width=0.85\linewidth]{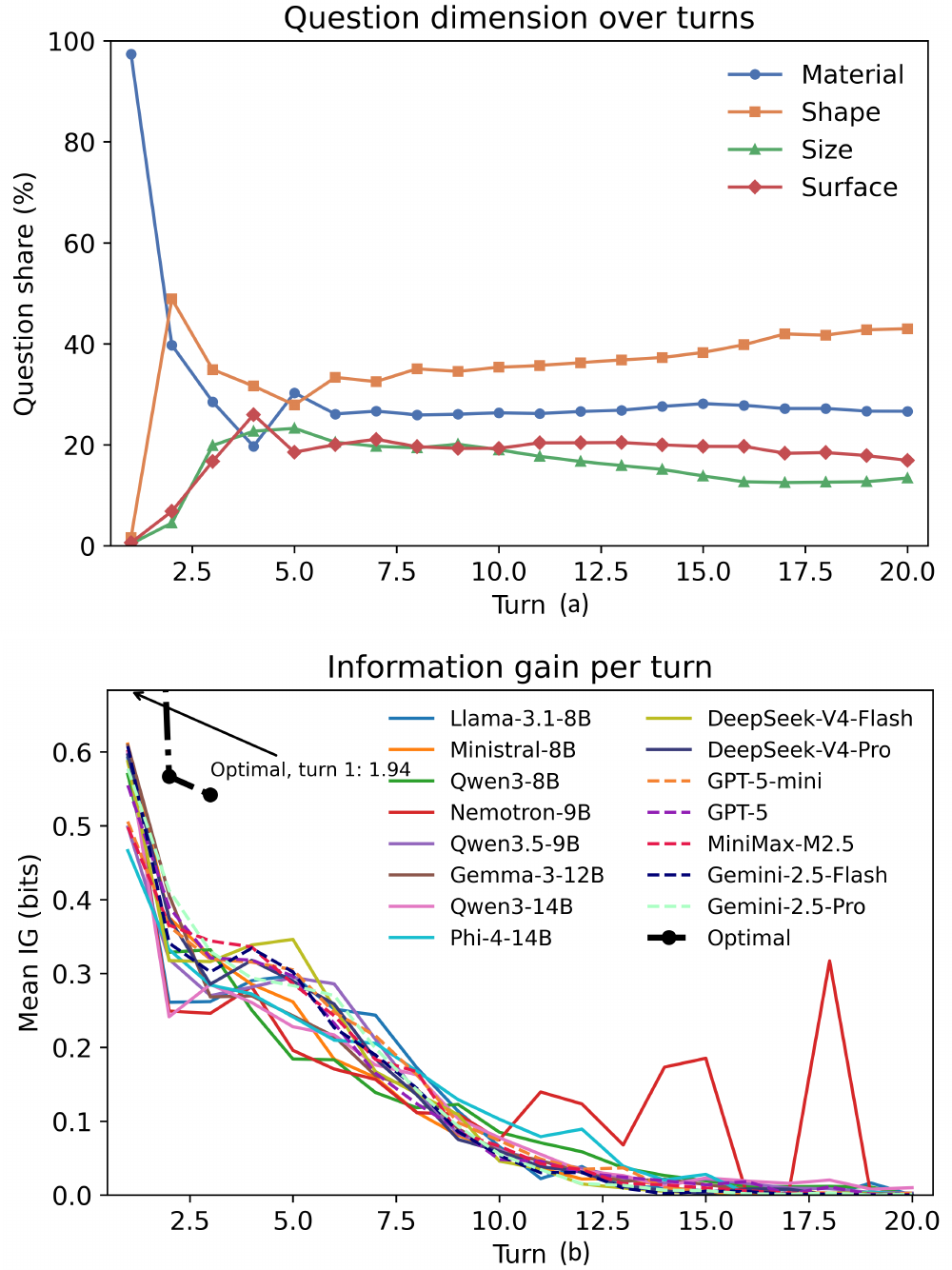}
\caption{Question behavior~(a) and information gain~(b) over turns.}
\label{fig:IG}
\end{figure}

\subsection{Question behavior and Information Gain}
We use the four atomic predicates in our KARI rule system to categorize LLM questions into different physical-property dimensions and visualize their share at each turn in \Cref{fig:IG}(a). We observe a clear trend: LLMs prefer to start the game with a material question and later adopt shape questions to guide their decisions. Size and surface questions occupy only a small share in the initial stage and grow afterward, while shape questions come to dominate the share (around 40\%). We further examine whether this question behavior is meaningful by scoring each turn with its information gain~(IG)~(\Cref{fig:IG}(b)). We find that material and shape questions provide useful information gain in the initial turns (1 to 5), although the large gap from \textsc{Optimal} shows these are still not the best questions to ask. The IG then collapses after turn~5 and gradually approaches zero, even though models keep asking questions from different categories. This shows that, in the early turns, LLMs can ask basic questions that roughly build up likelihoods over the candidate affordances to assist affordance reasoning. As the game proceeds, they cannot ask more discriminating questions needed to separate the target affordance from the remaining distractors, leaving them uncertain at the final guess. This also confirms our earlier observation that asking more questions does not improve the success rate.

\begin{figure}
\centering
\includegraphics[width=\linewidth]{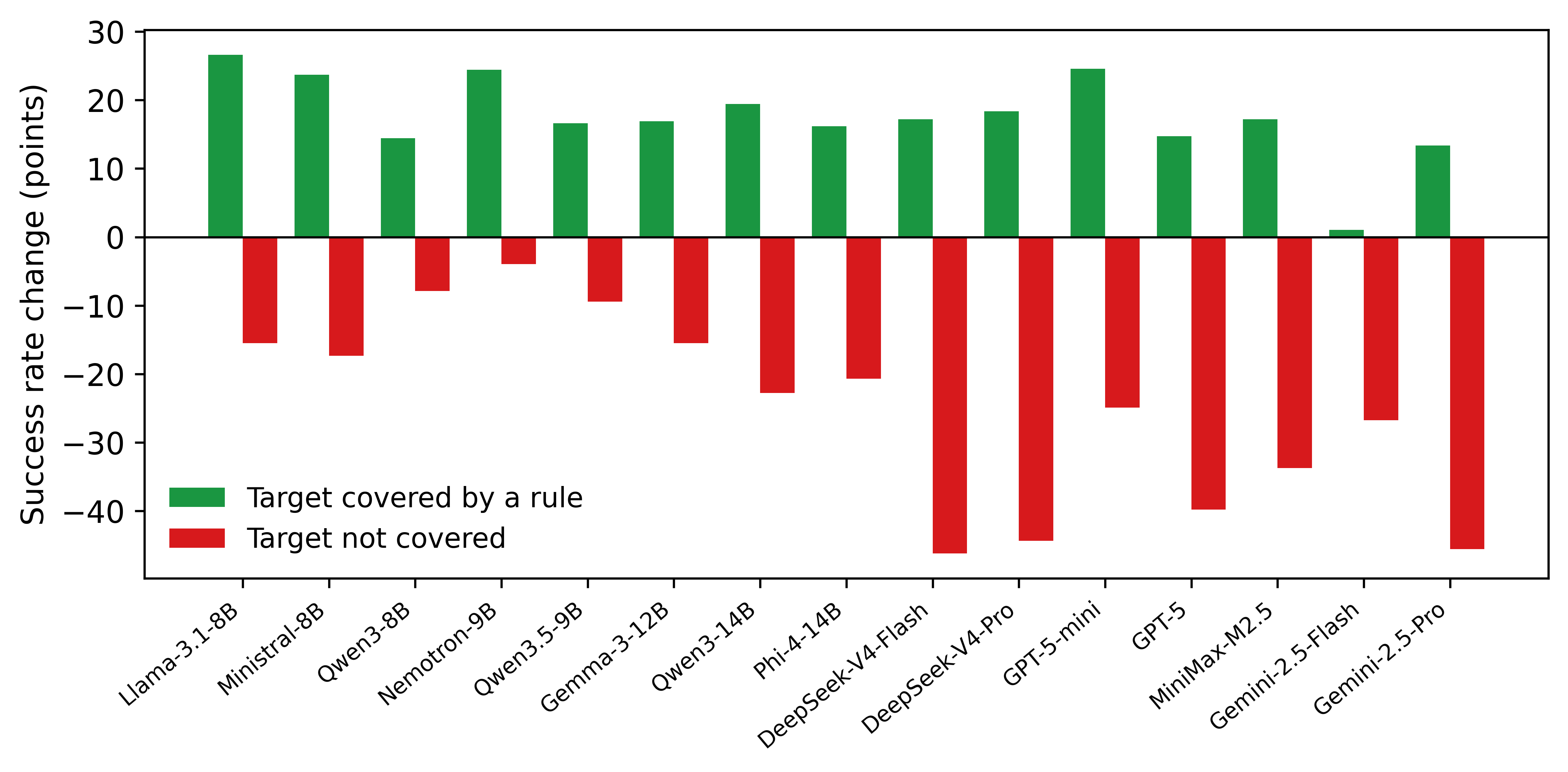}
\caption{Accuracy change split by the target affordance coverage.}
\label{fig:coverage}
\end{figure}

\subsection{KARI's Rule Integration}
\label{sec:results:kari}
The result of KARI's rule integration~(\Cref{tab:main}) is mainly two-fold. For models with 8B to 14B parameters, we observe a clear improvement in success rate, ranging from 4.1 to 15.2 points, verifying the effectiveness of the generated rules. We also notice that the rules have a mixed impact on the turn number for these models: a huge decrease is observed for \textsc{Qwen3-8B/14B}, while \textsc{Nemotron-9B} doubles its turns, and \textsc{Llama-3.1-8B} even uses all 20 turns. For most closed-source LLMs and DeepSeek-V4 series, we find that KARI's rules bring a drop in both success rate and turns, suggesting the rules lead these LLMs to make fast but unsure final guesses. KARI's rules are generated from knowledge bases whose limited coverage is a known limitation in prior work~\cite{bosselut2019comet,hwang2021comet}. We note that only 67.4\% of the games in \dataset~have their target affordance covered by a KARI rule. We split games by this coverage to analyze whether the limited coverage is a barrier to the improvement brought by KARI's rules~(\Cref{fig:coverage}). On covered target affordances, KARI's rules actually supply useful affordance knowledge and improve every LLM by 17.7 points on average. On uncovered ones, the distractor affordances' rules mislead the LLM toward irrelevant properties and lower the success rate by 25.0 points, hurting the strongest LLMs most, which can already solve some of these games even without rules~(further analysis in \Cref{app:kari_harm}).

\subsection{Case Study}
\label{sec:results:case}

\begin{figure}[t]                                         
\centering
\includegraphics[width=1\linewidth]{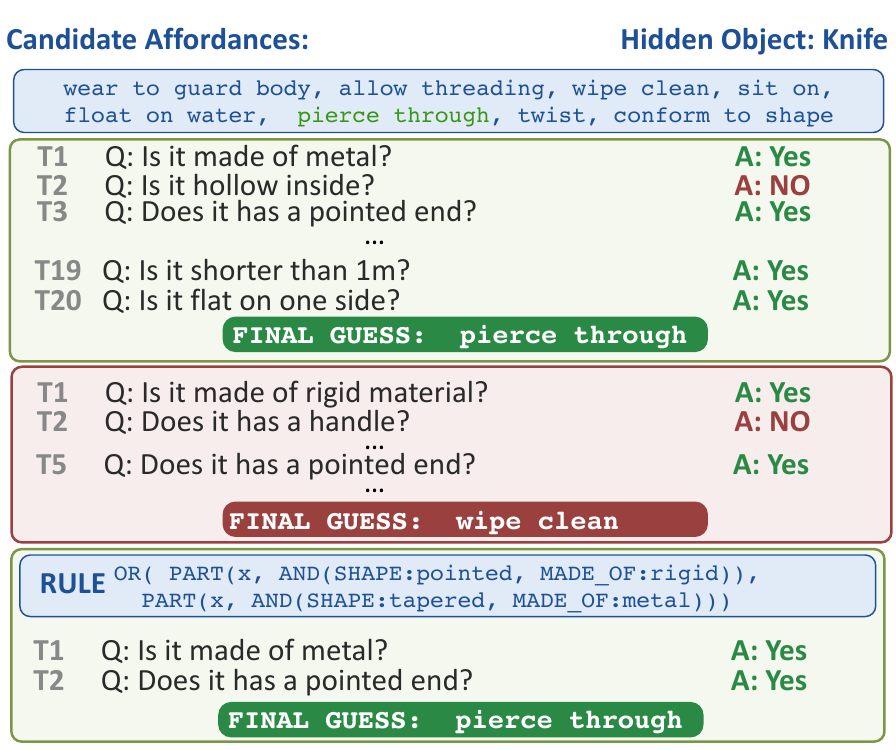}
\caption{Success and failure games with the target affordance \textit{pierce through} (hidden object: knife).}                                                                 
\label{fig:example}                                                                    
\end{figure} 
We present three games with the same game setting in \Cref{fig:example}. The first two use \textsc{DeepSeek-V4-Flash} and \textsc{MiniMax-M2.5} as the Questioner; the third reuses \textsc{DeepSeek-V4-Flash} assisted with KARI's rules. \footnote{In the actual game, rules mapped to the other candidate affordances are also presented. We omit them here for space.} In the first example, the model has already collected enough physical properties to support the target affordance \textit{pierce through} early in the game (T1, T3), yet it continues asking until the turn budget is exhausted. Although it eventually produces the correct final guess, the behavior suggests overthinking or low confidence in reasoning from physical properties to a specific affordance. The second example further illustrates this pattern: the model gathers sufficient evidence for the target affordance (T1, T5) but still arrives at a wrong final guess. In contrast, the third example shows that KARI's rule can provide explicit and effective affordance knowledge, allowing the same model to solve the game accurately in just two turns.

\section{Benchmark Validity}
\label{sec:validity}

The results above show a large gap between LLMs and humans. We further verify that this gap reflects affordance reasoning from physical properties rather than recall of object identity.

\begin{figure}[t]
\centering
\includegraphics[width=\linewidth]{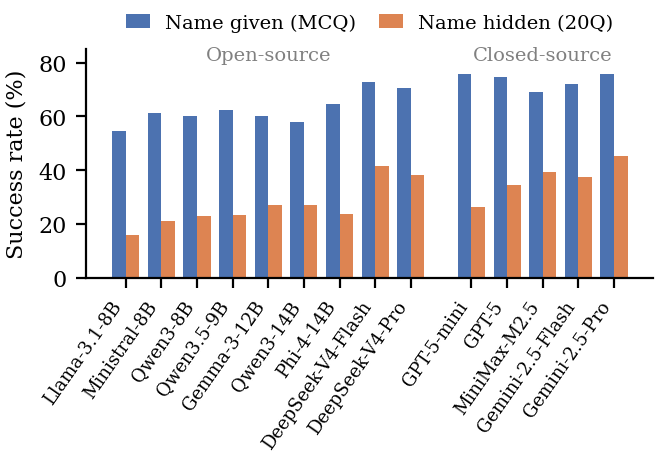}
\caption{Success rate on the same 295 games with the object name given (single-turn multiple choice) versus hidden (20Q).}
\label{fig:name_given}
\end{figure}

\paragraph{Existence of the recall shortcut.} We convert the games into single-turn multiple-choice questions where the object name is directly given and the model picks one of the 8 candidate affordances (5 games per target affordance, 295 games in total). As shown in \Cref{fig:name_given}, all models perform reasonably well (54.6\% to 75.6\%) once the name is given, far above their success rates on the same games with the name hidden. Models can answer from the object name alone, without exploring the object's physical properties, which is exactly the shortcut our hidden-identity design cuts off.

\paragraph{Identity remains hidden during the game.} A remaining concern is that models might still infer the object identity from the collected properties and then recall its affordance. For example, ``sharp, thin, made of metal'' suggests a knife, which affords cutting. We test this with a self-probe: for four Questioners, we sample 200 games that each model itself solved, give the model its own Q\&A transcript, and ask it to name the hidden object. Even on games they solved, models fail to name the object in around 90\% of them (\textsc{Qwen3-14B} 3.5\%, \textsc{DeepSeek-V4-Pro} 5.5\%, \textsc{GPT-5} 9.2\%, \textsc{Gemini-2.5-Pro} 8.5\%). Although each answer reveals one physical property, identifying the affordance requires far fewer properties than identifying the object, so the game usually ends before enough information accumulates to reveal the identity.

\paragraph{Robustness to counterintuitive properties.} We further modify the object properties in 30 games (e.g., a bladed tool's material is rewritten as soft and flexible) so that one original distractor becomes the new target, while the original target remains among the candidates as a tempting distractor. We run the same 20Q game on these modified games and compare with their original versions~(\Cref{tab:counterintuitive}). The models perform comparably on the modified games and rarely pick the tempting distractor, showing that our game structure ensures models collect physical properties through questioning and reasoning over them, rather than recalling the affordance of a typical object. We note that the number of feasible games is small, since \dataset~was not designed for this purpose, and a more comprehensive study requires collecting suitable data, which we leave to future work.

\begin{table}[t]
\centering
\small
\setlength{\tabcolsep}{3pt}
\begin{tabular}{lccc}
\toprule
& \multicolumn{2}{c}{SR (\%)} & Tempted (\%) \\
\cmidrule(lr){2-3}\cmidrule(lr){4-4}
Questioner & Original & Modified & Modified \\
\midrule
\textsc{Qwen3-14B}       & 20.0 & 50.0 &  0.0 \\
\textsc{DeepSeek-V4-Pro} & 36.7 & 30.0 & 16.7 \\
\textsc{GPT-5}           & 43.3 & 30.0 &  6.7 \\
\textsc{Gemini-2.5-Pro}  & 43.3 & 50.0 &  0.0 \\
\bottomrule
\end{tabular}
\caption{Results on the 30 games with counterintuitive properties. SR is the success rate on the original and the modified games. Tempted is the share of modified games in which the model picks the tempting distractor as its final guess.}
\label{tab:counterintuitive}
\end{table}
\section{Conclusion}

We introduced \dataset, a 20-Questions benchmark that measures affordance reasoning from physical properties rather than object-identity recall, comprising 1{,}009 games over 454 objects and 59 affordances. Our experiments with 15 state-of-the-art LLMs reveal a substantial gap~($\sim$20 points) compared to human performance, and a KL-based information-gain analysis shows that models fail to ask discriminating questions as the game progresses. To close the gap, we proposed KARI, a pipeline based on LLMs that generates affordance rules grounded in evidence from knowledge bases. KARI improves open-source LLMs by up to 15.2 points, while the limited coverage of KBs hinders further gains. We release the benchmark and all code, and hope \dataset~drives progress on physical reasoning that generalizes beyond memorized object-affordance mappings.
\newpage

\section{Limitations}

\paragraph{Scope of physical affordances.} \dataset~covers 59 affordances and 454 objects sampled from CSKG. We filter out affordances not deducible from physical-property dimensions alone, since their mechanisms cannot be reached by any material or shape question. For example, an induction cooktop heats via internal circuitry. Although some benchmarks~\cite{wang2026affordance} make initial attempts in this direction, a systematic study from non-physical properties to such affordances remains unexplored. 

\paragraph{Text-only domain.}
\dataset~is the first work to test a model's affordance reasoning ability from physical properties without exposing the object's identity. However, the whole benchmark is set up in a text-only setting. Exploring a similar setting in the visual domain can provide more insights into how current MLLMs perform on affordance reasoning, which also aligns with many real-world scenario tasks~\cite{jiang2026videop2r, driess2023palm,zhang2025landscape, jiang2024marvel}.

\paragraph{Question-Answering Setting.}
Our benchmark formulates affordance reasoning in a question-answering~(QA) setting. Real-world applications often require models to reason about affordances embedded in richer contexts, such as narrative understanding~\cite{jiang2023transferring, kovcisky2018narrativeqa}, open-ended planning~\cite{wang2023voyager} or long-horizon memories~\cite{liu2026conmem}, where affordance cues are implicit and must be inferred from context rather than elicited through explicit queries. Extending \dataset~to these settings can be a promising direction for future work.

\paragraph{Outcome-based evaluation.} Our main metric, success rate, only checks the final guess, while the 20Q game naturally has a step-wise structure. Prior work has shown that final accuracy alone can be unreliable for evaluating reasoning~\cite{sun2026beyond, fan2026chains,sheng2026solving}, since a model can reach the correct answer through inconsistent or hallucinated intermediate steps. Our information-gain analysis is a first step toward evaluating the process, and future work can build process-level metrics on the game transcripts to evaluate the hallucination and consistency of models' reasoning.

\section{Ethical Considerations }

\paragraph{Data sources.}
Our affordance vocabulary and positive evidence are drawn from CSKG~\cite{ilievski2021cskg}, which is publicly released for research use. The benchmark contains only physical-property descriptions of everyday objects and includes no personal or sensitive data.

\paragraph{Intended use and potential risks.}
\dataset~is intended for research on physical affordance reasoning.  While prior work has shown
can be exploited to elicit harmful behaviors from LLMs through
multi-turn attacks~\cite{jiang2025red, wang2025comprehensive}, our task is strictly confined
to querying physical properties of everyday objects, and thus the benchmark and KARI rules contain no offensive content. We note that closed-source LLM scores reported here may shift as providers update their underlying models.
\section*{Acknowledgments} This material is based upon work supported by the Defense Advanced Research Projects Agency (DARPA) under Agreement No. HR00112590089. This research was sponsored by the Defense Advanced Research Projects Agency via Contract HR00112390061.
\bibliography{custom}

\newpage
\appendix
\section{KARI Pipeline}
\label{app:kari_pipe}
\subsection{Prompts}
\label{app:kari:prompts}

We list the system prompts for three components as follows:

\paragraph{Proposer.}
The proposer writes the rule given the affordance verbs and corresponding positive objects with physical properties from KB.

\begin{lstlisting}[basicstyle=\ttfamily\footnotesize, breaklines=true, frame=single, backgroundcolor=\color{black!2}, columns=fullflexible]
You are a physical reasoning expert. Given an affordance verb and
example objects that can perform it, produce a rule describing the
PHYSICAL PREREQUISITES.

## GRAMMAR (output valid JSON only)

Logic (2) -- combine sub-expressions:
  {"op": "AND", "args": [expr, ...]}
  {"op": "OR",  "args": [expr, ...]}

Constructor (1) -- introduce a named part of the object:
  {"op": "PART", "role": "<var>", "body": expr}

Atomic predicates (4) -- attach a static physical property:
  {"op": "MADE_OF",  "role": "<whole|var>", "class": "<material>"}
  {"op": "SHAPE",    "role": "<whole|var>", "class": "<shape>"}
  {"op": "SIZE",     "role": "<whole|var>", "class": "<size>"}
  {"op": "SURFACE",  "role": "<whole|var>", "class": "<surface>"}

## ROLE SEMANTICS
- "whole" = the entire object.
- "x" / "y" / "z" = named parts introduced by PART.

## CLASS VALUES
Use concise canonical terms describing STATIC PHYSICAL structure
(composition, geometry, size, surface texture).

## HARD RULES
- NEVER use affordance verbs or their adjective forms
  (forbidden: "cuttable", "reflective", "absorbent").
- NEVER use dynamic behavior words
  (forbidden: "compressible", "foldable"); describe the static
  property that enables the behavior instead (e.g. "foam" not
  "compressible").
- Every PART must introduce a fresh variable name.
- Every atom's "role" must be "whole" or a previously declared
  PART variable (no free variables).

## OUTPUT
JSON rule only. No markdown, no prose.
\end{lstlisting}

\paragraph{Validator.} The validator labels each atom of the current rule \textit{YES}/\textit{NO} for a given object.

\begin{lstlisting}[basicstyle=\ttfamily\footnotesize, breaklines=true, frame=single, backgroundcolor=\color{black!2}, columns=fullflexible]
For each (op, class) atom, decide YES/NO: does the given object have
this STATIC PHYSICAL property, based on the raw evidence plus your
world knowledge of the object?

A PART-level atom (e.g. SHAPE=sharp_edge) holds if ANY part of the
object has the property -- not only the whole. E.g. a knife has
SHAPE=sharp_edge because its blade does.

## OUTPUT (JSON array)
[{"op": "<OP>", "class": "<val>", "holds": true|false}, ...]

## OUTPUT FORMAT
JSON array only. No markdown, no prose.
\end{lstlisting}

\paragraph{Auditor.} The auditor checks the value of each predicate and updates new ones into the vocabulary.

\begin{lstlisting}[basicstyle=\ttfamily\footnotesize, breaklines=true, frame=single, backgroundcolor=\color{black!2}, columns=fullflexible]
For each proposed class value (none of them literally equals any
vocab word -- that's been pre-filtered), decide:
- SYNONYM: near-exact semantic equivalence with a vocab word
           -> map to that word.
- NEW:     genuinely new concept, no clear synonym in vocab
           -> mapped_to = null.

## OP: {op_name}
## VOCABULARY: {vocab_for_this_op}

## OUTPUT (JSON array)
[{"proposed": "<val>", "decision": "SYNONYM|NEW",
  "mapped_to": "<vocab_word>|null"}, ...]

## HARD RULE
SYNONYM requires the SAME underlying concept within this op's
semantic dimension. When unsure, output NEW.

## OUTPUT FORMAT
JSON array only. No markdown, no prose.
\end{lstlisting}

\subsection{Rule Examples}
\label{app:kari:rules}

We present two rules to illustrate KARI's tree structure. \textit{pierce\_through} (\Cref{lst:rule_pierce}) shows three alternative physical pathways, each scoped to a specific part. \textit{fold\_flat} (\Cref{lst:rule_fold}) shows four alternatives that all describe the whole object.

\begin{lstlisting}[basicstyle=\ttfamily\footnotesize, breaklines=true, frame=single, backgroundcolor=\color{black!2}, columns=fullflexible, caption={Rule for \textit{pierce\_through}.}, label={lst:rule_pierce}]
OR(
  PART(x, AND(SHAPE: pointed,     MADE_OF: rigid_material, SIZE: thin)),
  PART(x, AND(SHAPE: tapered,     MADE_OF: metal)),
  PART(x, AND(SHAPE: cylindrical, MADE_OF: metal,          SIZE: long))
)
\end{lstlisting}

\begin{lstlisting}[basicstyle=\ttfamily\footnotesize, breaklines=true, frame=single, backgroundcolor=\color{black!2}, columns=fullflexible, caption={Rule for \textit{fold\_flat}.}, label={lst:rule_fold}]
OR(
  AND(MADE_OF: flexible_material, SHAPE: flat,        SURFACE: smooth),
  AND(MADE_OF: flexible_material, SHAPE: rectangular, SIZE:    thin),
  AND(MADE_OF: flexible_material, SHAPE: cylindrical),
  AND(MADE_OF: flexible_material, SHAPE: curved,      SIZE:    broad)
)
\end{lstlisting}
\section{Further Analysis of KARI's Rule Integration}
\label{app:kari_harm}

We further explore what makes rule injection harmful. We first measure how often each model's final guess falls on a candidate that has a rule~(\Cref{tab:kari_covered_guess}). On average, 50.6\% of the candidates in each game carry a rule, so a random guess lands on a rule-covered candidate with this probability. The vanilla setting stays close to this level (55--57\%), showing that models do not prefer rule-covered candidates on their own. With rules injected, the share jumps to 84--87\%, showing that rule injection changes the models' choice preference toward rule-covered candidates.

\begin{table}[h]
\centering
\small
\begin{tabular}{lcc}
\toprule
Guess on a rule-covered candidate & Vanilla & +KARI \\
\midrule
Open-source LLMs   & 56.8\% & 83.7\% \\
Closed-source LLMs & 55.3\% & 87.2\% \\
\bottomrule
\end{tabular}
\caption{Share of final guesses that fall on a rule-covered candidate.}
\label{tab:kari_covered_guess}
\end{table}

We then examine games where the target and distractors are both covered by rules, grouped by the number of rule-covered distractors in the candidate list~(\Cref{tab:kari_gradient}). Groups with 0 and 6 covered distractors are omitted due to small sample sizes. The gain shrinks as more distractors also carry rules, but the rules remain helpful, so the drop does not come from conflicts between the rules and the models' internal knowledge. In summary, rule injection shifts the models' choice toward rule-covered candidates, so models can be misled on games they would have solved without rules. This effect is amplified for models with stronger vanilla performance, since they have more correct answers to lose, and eventually leads to their overall drop.

\begin{table}[h]
\centering
\small
\setlength{\tabcolsep}{6pt}
\begin{tabular}{ccc}
\toprule
\# Rule-covered & \multicolumn{2}{c}{$\Delta$ SR (\%)} \\
\cmidrule(lr){2-3}
distractors & Open-source & Closed-source \\
\midrule
1 & +34.9 & +31.2 \\
2 & +32.5 & +27.8 \\
3 & +12.9 & +6.1 \\
4 & +17.0 & +11.5 \\
5 & +11.8 & +7.5 \\
\bottomrule
\end{tabular}
\caption{Success rate change from KARI's rules on covered targets, by number of rule-covered distractors.}
\label{tab:kari_gradient}
\end{table}
\section{Experiment Setup Details}
\label{app:setup}

\subsection{Fix20Q Baseline Question Selection}
\label{app:fixedq}
The Fixed-Q baseline asks 20 yes/no questions in a fixed order, curated from question produced by all evaluated LLM questioners on the test split~(\Cref{tab:fixedq}). As \Cref{tab:fixedq} shows, the questions asked by LLMs are open-vocabulary and follow a long-tail distribution: the 20 most frequent questions account for only 10.2\% of all 373,060 yes/no turns. A deterministic lookup oracle can only answer a predefined question set, which would limit the Questioners' questioning ability, so we adopt an LLM Oracle grounded on the property set, as in prior 20Q work~\cite{bertolazzi2023chatgpt,hutson2025guessinggame}.

\subsection{Entropy-Max Baseline}
\label{app:entropy}
\textsc{Entropy-Max} uses the same representative objects and stopping threshold~(0.9) as \textsc{Optimal}, but has no access to the target. It reaches 51.2\% in 4.6 turns, outperforming all LLMs in \Cref{tab:main} but still falling far short of \textsc{Optimal}. Since its questions never aim at any target, its likelihood often converges on a wrong candidate, showing that narrowing the search space alone is not enough for success.

\begin{table*}[t]
\centering
\small
\setlength{\tabcolsep}{4pt}
\begin{tabular}{rllrp{0.42\linewidth}}
\toprule
\# & Dimension & Property & Count (\%) & Question \\
\midrule
 1 & Material & rigid        & 28107 (7.5) & Is the object made of a rigid material? \\
 2 & Surface  & smooth       & 21613 (5.8) & Is the object's surface smooth? \\
 3 & Material & metal        & 18816 (5.0) & Is the object made of metal? \\
 4 & Material & flexible     & 18248 (4.9) & Is the object made of a flexible material? \\
 5 & Shape    & hollow       & 17952 (4.8) & Is the object hollow? \\
 6 & Material & fabric       & 13436 (3.6) & Is the object made of fabric? \\
 7 & Shape    & cylindrical  & 11157 (3.0) & Is the object's shape cylindrical? \\
 8 & Shape    & hollow  & 10068 (2.7) & Is the object hollow inside? \\
 9 & Material & plastic      &  9442 (2.5) & Is the object made of plastic? \\
10 & Shape    & flat         &  7172 (1.9) & Is the object flat? \\
11 & Material & wood         &  6375 (1.7) & Is the object made of wood? \\
12 & Material & fibrous      &  6047 (1.6) & Is the object made of a fibrous material? \\
13 & Shape    & elongated    &  5969 (1.6) & Is the object's shape elongated? \\
14 & Size     & thin         &  5831 (1.6) & Is the object thin? \\
15 & Shape    & rectangular  &  5593 (1.5) & Is the object's shape rectangular? \\
16 & Surface  & rough        &  5500 (1.5) & Is the object's surface rough? \\
17 & Material & glass        &  5026 (1.3) & Is the object made of glass? \\
18 & Shape    & layered      &  4159 (1.1) & Is the object layered? \\
19 & Shape    & bowl-shaped  &  4123 (1.1) & Is the object's shape bowl-shaped? \\
20 & Shape    & solid        &  4033 (1.1) & Is the object solid? \\
\bottomrule
\end{tabular}
\caption{The 20 most frequent physical-property questions across all evaluation games, with their targeted dimension and count (percentage of the 373{,}060 yes/no turns).}
\label{tab:fixedq}
\end{table*}

\subsection{Prompts}
\label{app:20Q:prompts}

\paragraph{Questioner.}
The Questioner plays the active role of the game and asks yes/no questions about the hidden object's static physical properties.

\begin{lstlisting}[basicstyle=\ttfamily\footnotesize, breaklines=true, frame=single, backgroundcolor=\color{black!2}, columns=fullflexible]
You are playing a 20 Questions game about object AFFORDANCES.

One of the following affordance labels describes the hidden object.
Your goal is to identify it by asking yes/no questions about the
object's static physical properties.

Candidate affordances:
  1. <affordance_1>
  2. <affordance_2>
  ...

Ask questions along these FOUR physical dimensions:
  - MATERIAL -- what it is made of (composition, rigidity/flexibility,
                fibrousness)
  - SHAPE    -- geometric outline, parts, OR topology
                (hollow / solid / enclosed / layered)
  - SIZE     -- dimensions (how big / long / thick)
  - SURFACE  -- outer surface texture

Rules:
- Ask ONE yes/no question per turn, about EXACTLY ONE property from
  EXACTLY ONE dimension above.
- Do NOT ask directly which affordance it is.
- Do NOT ask about the object's function, use, purpose, activity,
  or category.

BAD questions to AVOID:
  function / use / category:
    "Is it used for cutting?"
    "Does it absorb liquid?"
    "Is it a tool?"
  multiple properties at once (non-atomic):
    "Is it both metal and pointed?"
    "Is it flexible and smooth?"

When you are confident (or after 20 questions), output your final
guess using EXACTLY this format on its own line:
  FINAL_GUESS: <affordance label>
  where <affordance label> must be one of the candidates above.

Output format each turn:
QUESTION: <your yes/no question>
\end{lstlisting}

\paragraph{Oracle.}
The Oracle sees the full description of the hidden object and replies with a single yes/no per question.

\begin{lstlisting}[basicstyle=\ttfamily\footnotesize, breaklines=true, frame=single, backgroundcolor=\color{black!2}, columns=fullflexible]
You are the Oracle in a 20 Questions game about physical objects.
You know the hidden object described below. Your task is to answer
yes/no questions about the object's physical properties -- shape,
material, size, surface, structural parts, and other observable
physical characteristics.

Rules:
- Answer ONLY with "Yes" or "No" (one word).
- The description below is provided as supplementary context to
  identify which object you are reasoning about.
- Do NOT reveal the object's name.

Hidden object description:
<object_description>
\end{lstlisting}

\paragraph{Checker.}
The Checker classifies each question into one or more \texttt{DIMENSION:value} pairs and rejects questions that probe function, use, or category. Output is one pair per line, used by the game loop to enforce question atomicity.

\begin{lstlisting}[basicstyle=\ttfamily\footnotesize, breaklines=true, frame=single, backgroundcolor=\color{black!2}, columns=fullflexible]
Classify the question and list EVERY DIMENSION:value pair the
question EXPLICITLY probes, one per line. Do NOT infer, expand,
paraphrase, or add synonyms -- output only what the question
literally mentions, using the question's own wording for the value.

Dimensions (with one example each):
  MATERIAL  - composition, rigidity/flexibility, fibrousness
              "Is it made of metal?"  -> MATERIAL:metal
              "Is it rigid?"          -> MATERIAL:rigid
  SHAPE     - geometric outline, the existence of a geometric part,
              OR internal/whole-body topology (hollow, solid,
              enclosed, layered, ...).
              "Is it cylindrical?"         -> SHAPE:cylindrical
              "Does it have a sharp edge?" -> SHAPE:sharp_edge
              "Is it hollow?"              -> SHAPE:hollow
  SIZE      - dimension (how big / long / thick / small)
              "Is it pocket-sized?" -> SIZE:pocket_sized
  SURFACE   - outer surface texture
              "Is it smooth?" -> SURFACE:smooth
  FUNCTION  - use / purpose / activity / what the object does
              "Is it used for cutting?" -> FUNCTION:cutting
  CATEGORY  - class / type / what kind of object it is
              "Is it a tool?" -> CATEGORY:tool

Comparative SIZE -- preserve the comparison direction and reference
object. Do NOT collapse a comparison into a canonical size word.
  "Is it larger than a hand?"  -> SIZE:larger_than_hand
  "Is it smaller than a coin?" -> SIZE:smaller_than_coin

Single vs multi mapping:
  "Is it long?"            -> SIZE:long
  "Is it long and narrow?" -> SIZE:long
                              SIZE:narrow
  "Is it made of metal and shiny?" -> MATERIAL:metal
                                      SURFACE:shiny

Output one DIMENSION:value pair per line. No extra text.
\end{lstlisting}
\subsection{Implementation Details}
\label{app:impl}

We evaluate 15 LLM questioners. Open-source models (Llama-3.1-8B, Ministral-8B, Qwen3-8B/14B, Qwen3.5-9B, Nemotron-9B, Gemma-3-12B, Phi-4-14B) are served locally with sglang on 8\,$\times$\,NVIDIA RTX A6000 GPUs (48\,GB each). Closed-source models (\textsc{GPT-5}, \textsc{GPT-5-mini}, \textsc{Gemini-2.5-Flash}, \textsc{Gemini-2.5-Pro}, \textsc{MiniMax-M2.5}) and the two DeepSeek-V4 variants are accessed through commercial APIs (OpenAI, Google AI Studio, OpenRouter). All decoding uses temperature 0, with a per-call \texttt{max\_tokens} budget of 512 for the Questioner, 20 for the Oracle, and 50 for the Checker. The Oracle and the Checker are both instances of \textsc{Qwen3-14B} served locally by the same sglang backend. Each game runs to a hard cap of 20 turns, after which a forced guess is requested. If the Checker rejects three consecutive questions, the game is terminated and counted as a failure.

\Cref{tab:qc_terminated} reports the share of games terminated by the Checker for each Questioner. Most models are affected in fewer than 4\% of games. \textsc{GPT-5-mini} (36.4\%) and \textsc{Qwen3.5-9B} (26.4\%) are the exceptions, and excluding their terminated games raises their success rates to 35.8\% and 34.1\%, but the overall ranking and the gap to human performance remain unchanged, so the game format is not the main bottleneck.

\begin{table}[h]
\centering
\small
\setlength{\tabcolsep}{3pt}
\begin{tabular}{lccc}
\toprule
Questioner & Terminated (\%) & SR & SR excl. \\
\midrule
\textsc{Llama-3.1-8B}      &  1.5 & 18.4 & 18.7 \\
\textsc{Ministral-8B}      & 15.8 & 17.7 & 21.1 \\
\textsc{Qwen3-8B}          &  2.2 & 26.3 & 26.8 \\
\textsc{Nemotron-9B}       & 13.2 & 14.9 & 17.1 \\
\textsc{Qwen3.5-9B}        & 26.4 & 25.1 & 34.1 \\
\textsc{Gemma-3-12B}       &  0.0 & 27.4 & 27.4 \\
\textsc{Qwen3-14B}         &  0.2 & 27.3 & 27.3 \\
\textsc{Phi-4-14B}         &  3.8 & 26.4 & 27.4 \\
\textsc{DeepSeek-V4-Flash} &  1.0 & 41.3 & 41.7 \\
\textsc{DeepSeek-V4-Pro}   &  0.9 & 40.5 & 40.9 \\
\midrule
\textsc{GPT-5-mini}        & 36.4 & 22.8 & 35.8 \\
\textsc{GPT-5}             &  3.8 & 35.5 & 36.9 \\
\textsc{MiniMax-M2.5}      & 10.2 & 34.9 & 38.9 \\
\textsc{Gemini-2.5-Flash}  &  6.7 & 41.4 & 44.4 \\
\textsc{Gemini-2.5-Pro}    &  3.4 & 45.9 & 47.5 \\
\bottomrule
\end{tabular}
\caption{Share of games terminated by the Checker, with success rate (\%) before and after excluding them.}
\label{tab:qc_terminated}
\end{table}
\subsection{Oracle Validation}
\label{app:oracle}

To verify that the Oracle is accurate enough to drive the game loop, we collect 300 yes/no questions sampled from the test set and have three human annotators label the ground-truth answer for each. We then run three candidate Oracle models on the same set and report agreement with the human labels in \Cref{tab:oracle_validation}.

We further break down the agreement of \textsc{Qwen3-14B} by property dimension in \Cref{tab:oracle_breadown}. Errors concentrate in SIZE questions that compare the object against an external reference (e.g., ``Is the object larger than a shoebox?''), where the answer is not directly stated in the property set.

To test whether the Oracle choice affects the results, we replace \textsc{Qwen3-14B} with \textsc{GPT-5} as the Oracle and re-run six Questioners on all 1,009 games~(\Cref{tab:oracle_gpt5}). Success rates shift by at most 2.6 points~(mean 1.5), the ranking is unchanged, and the direction of effect of KARI's rules is identical.

\begin{table*}[h]
\centering
\setlength{\tabcolsep}{4pt}
\begin{tabular}{lcccc}
\toprule
& \multicolumn{2}{c}{Vanilla} & \multicolumn{2}{c}{+KARI} \\
\cmidrule(lr){2-3}\cmidrule(lr){4-5}
Questioner & \textsc{Qwen3-14B} & \textsc{GPT-5} & \textsc{Qwen3-14B} & \textsc{GPT-5} \\
\midrule
\textsc{Qwen3.5-9B}   & 25.1 & 23.3 & 33.2 & 32.6 \\
\textsc{Qwen3-14B}    & 27.3 & 24.7 & 32.9 & 34.5 \\
\textsc{Gemma-3-12B}  & 27.4 & 27.9 & 33.7 & 34.7 \\
\textsc{GPT-5-mini}   & 22.8 & 24.4 & 31.2 & 31.8 \\
\textsc{GPT-5}        & 35.5 & 37.9 & 32.4 & 34.8 \\
\textsc{MiniMax-M2.5} & 34.9 & 35.2 & 35.5 & 33.7 \\
\bottomrule
\end{tabular}
\caption{Success rate (\%) with \textsc{Qwen3-14B} and \textsc{GPT-5} as the Oracle.}
\label{tab:oracle_gpt5}
\end{table*}

\begin{table}[t]
\centering
\small
\setlength{\tabcolsep}{6pt}
\begin{tabular}{lccc}
\toprule
 & \textsc{Qwen3-8B} & \textsc{Qwen3-14B} & \textsc{GPT-5} \\
\midrule
Accuracy (\%)      & 85.2   & 93.4   & 95.5   \\
\bottomrule
\end{tabular}
\caption{Oracle agreement with human labels on 300 sampled questions.}
\label{tab:oracle_validation}
\end{table}

\begin{table}[t]
\centering
\small
\setlength{\tabcolsep}{3pt}
\begin{tabular}{lcccc}
\toprule
\textsc{Qwen3-14B} & MATERIAL & SHAPE & SURFACE & SIZE \\
\midrule
Accuracy (\%) & 98.0 & 95.7 & 88.2 & 84.8 \\
\bottomrule
\end{tabular}
\caption{Oracle agreement with human labels by property dimension.}
\label{tab:oracle_breadown}
\end{table}
\section{Success Rate Breakdown}
\label{app:sr_breakdown}

We report the top-5 easiest and bottom-5 hardest affordances for three model groups: 1)~All 15 LLMs~(\Cref{tab:sr_avg}), 2)~Open-source LLMs~(\Cref{tab:sr_open5}), and 3)~Closed-source LLMs~(\Cref{tab:sr_closed5}).

\begin{table}[t]
\centering
\footnotesize
\setlength{\tabcolsep}{6pt}
\begin{tabular}{lr}
\toprule
Affordance & SR \\
\midrule
\multicolumn{2}{l}{\textit{Top 5}} \\
abrade\_surface           & 57.8 \\
conduct\_heat & 55.2 \\
enclose\_substance        & 53.3 \\
store\_liquid             & 51.9 \\
transmit\_light           & 51.4 \\
\midrule
\multicolumn{2}{l}{\textit{Bottom 5}} \\
hang\_from\_above         & 10.6 \\
float\_on\_water          &  7.2 \\
hold\_between             &  6.7 \\
ignite                    &  4.9 \\
sink\_in\_water           &  3.8 \\
\bottomrule
\end{tabular}
\caption{Easiest and hardest affordances, averaged over all 15 vanilla questioners.}
\label{tab:sr_avg}
\end{table}

\begin{table}[t]
\centering
\footnotesize
\setlength{\tabcolsep}{6pt}
\begin{tabular}{lr}
\toprule
Affordance & SR \\
\midrule
\multicolumn{2}{l}{\textit{Top 5}} \\
abrade\_surface           & 73.3 \\
store\_liquid             & 70.4 \\
transmit\_light           & 65.7 \\
conduct\_heat & 63.8 \\
enclose\_substance        & 60.9 \\
\midrule
\multicolumn{2}{l}{\textit{Bottom 5}} \\
hang\_from\_above         &  9.1 \\
float\_on\_water          &  8.7 \\
ignite                    &  4.3 \\
hold\_between             &  2.9 \\
sink\_in\_water           &  2.6 \\
\bottomrule
\end{tabular}
\caption{Easiest and hardest affordances for the top-5 open-source models (\textsc{DeepSeek-V4-Flash}, \textsc{DeepSeek-V4-Pro}, \textsc{Gemma-3-12B}, \textsc{Qwen3-14B}, \textsc{Phi-4-14B}).}
\label{tab:sr_open5}
\end{table}

\begin{table}[t]
\centering
\footnotesize
\setlength{\tabcolsep}{6pt}
\begin{tabular}{lr}
\toprule
Affordance & SR \\
\midrule
\multicolumn{2}{l}{\textit{Top 5}} \\
bounce\_back              & 71.4 \\
enclose\_substance        & 70.4 \\
pierce\_through           & 62.6 \\
store\_liquid             & 61.7 \\
abrade\_surface           & 60.0 \\
\midrule
\multicolumn{2}{l}{\textit{Bottom 5}} \\
sweep                     & 13.3 \\
hold\_between             & 11.4 \\
ignite                    &  9.6 \\
float\_on\_water          &  7.8 \\
sink\_in\_water           &  7.0 \\
\bottomrule
\end{tabular}
\caption{Easiest and hardest affordances for the top-5 closed-source models (\textsc{Gemini-2.5-Pro}, \textsc{Gemini-2.5-Flash}, \textsc{GPT-5}, \textsc{MiniMax-M2.5}, \textsc{GPT-5-mini}).}
\label{tab:sr_closed5}
\end{table}
\section{Human Annotation}
\label{app:annotation}

\subsection{Participants and Instructions.}
We recruited annotators for four roles: 
\paragraph{(1)} Five volunteers who played the human baseline on the 30\% subset. The instructions are the same as Questioner~(\Cref{app:20Q:prompts}). 

During the games, we observe that participants are limited by the 20-turn budget. Around turn 10, when the collected information is still not decisive, they tend to pick among the 2--3 remaining candidates, and some participants report that they cannot come up with a proper question to further distinguish them.

\paragraph{(2)} Three annotators who verified the Oracle answers on a sampled 300-question subset reported in \Cref{tab:oracle_validation}. The instructions are the same as Oracle~(\Cref{app:20Q:prompts}). 

\paragraph{(3,4)} Six annotators were involved in our manual annotation and refinement process in Stages 2 and 3 of \dataset collection and annotation. Three additional annotators verified our labels for the object-affordance pair. ALL instructions as follows:
\begin{lstlisting}[basicstyle=\ttfamily\footnotesize, breaklines=true, frame=single, backgroundcolor=\color{black!2}, columns=fullflexible]
You are participating in an annotation task to construct structured
physical descriptions for everyday objects.

Your goal is to describe each object using only observable physical
attributes, including shape, material composition, surface
characteristics, dimensions, and structural parts.

Guidelines:
- Focus strictly on physical and geometric properties.
- Do not include functional, social, or usage-related descriptions.
- Materials should be written as material names only
  (e.g., metal, wood, plastic, stainless steel).
- Surface and shape attributes should describe only physical
  appearance or geometry.
- Part-specific properties should be assigned to the corresponding
  object parts rather than the global object description.
- The text description should contain 3--6 sentences describing
  only physical characteristics.
- Do not include phrases such as "used for", "designed to",
  or "can be used".
- Global size should reflect realistic dimensions in centimeters.

Please provide the final annotation in a JSON format.
\end{lstlisting}

\begin{lstlisting}[basicstyle=\ttfamily\footnotesize, breaklines=true, frame=single, backgroundcolor=\color{black!2}, columns=fullflexible]
You are participating in an annotation task for a physical affordance benchmark.

For each listed object, determine which affordances the object
physically supports based only on its observable properties,
including shape, geometry, material, surface characteristics,
and structural components.

An affordance should be assigned only if the object's physical
properties reasonably satisfy the affordance definition.

Guidelines:
- Base decisions only on visible or physically inferable properties.
- Do not consider electronic, chemical, or mechanism-specific
  functions unless they are directly implied by the object's
  physical structure.
- If an affordance clearly applies, include it; if uncertain,
  leave it unassigned.

Please provide annotations in the following JSON format:
[
  {"object": "<name>", "has": ["<affordance_name>", ...]},
  ...
]
\end{lstlisting}

\subsection{Recruitment and payment.}
All participants were university students who voluntarily participated through internal recruitment channels. Participants did not receive any financial compensation.

\subsection{Data Consent}
All participants were informed that their annotations would be used solely for academic research and dataset construction purposes prior to participation. And no demographic or personally identifying information was retained beyond the annotation outputs.

\end{document}